# Partnering Strategies for Fitness Evaluation in a Pyramidal Evolutionary Algorithm

**Uwe Aickelin**

School of Computer Science
University of Nottingham
NG8 1BB UK
uxa@cs.nott.ac.uk

**Larry Bull**

Intelligent Computer Systems Centre
University of the West of England
Bristol BS16 1QY, UK

## Abstract

This paper combines the idea of a hierarchical distributed genetic algorithm with different inter-agent partnering strategies. Cascading clusters of sub-populations are built from bottom up, with higher-level sub-populations optimising larger parts of the problem. Hence higher-level sub-populations search a larger search space with a lower resolution whilst lower-level sub-populations search a smaller search space with a higher resolution. The effects of different partner selection schemes for (sub-)fitness evaluation purposes are examined for two multiple-choice optimisation problems. It is shown that random partnering strategies perform best by providing better sampling and more diversity.

## 1 INTRODUCTION

When hierarchically distributed evolutionary algorithms are combined with multi-agent structures a number of new questions become apparent. One of these questions is addressed in this paper: the issue of assigning a meaningful (sub-) fitness to an agent. This paper will look at seven different partnering strategies for fitness evaluation when combined with a genetic algorithm that uses a co-operative sub-population structure. We will evaluate the different strategies according to their optimisation performance of two scheduling problems.

Genetic algorithms are generally attributed to Holland [1976] and his students in the 1970s, although evolutionary computation dates back further (refer to Fogel [1998] for an extensive review of early approaches). Genetic algorithms are stochastic meta-heuristics that mimic some features of natural evolution. Canonical genetic algorithms were not intended for function optimisation, as discussed by De Jong [1993]. However, slightly modified versions proved very successful. For an introduction to genetic algorithms for function optimisation, see Deb [1996].

The twist when applying our type of distributed genetic algorithm lies in its special hierarchical structure. All sub-populations follow different (sub-) fitness functions, so in effect only searching specific parts of the solution space. Following special crossover-operators these parts are then gradually merged to full solutions. The advantage of such a divide and conquer approach is reduced epistasis within the lower-level sub-populations which makes the optimisation task easier for the genetic algorithm.

The paper is arranged as follows: the following section describes the nurse scheduling and tenant selection problems. Pyramidal genetic algorithms and their application to these two problems are detailed in section 3. Section 4 explains the seven partnering strategies examined in the paper and section 5 describes their use and computational results. The final section discusses all findings and draws conclusions.

## 2 THE NURSE SCHEDULING PROBLEM

Two optimisation problems are considered in this paper, the nurse scheduling problem and the tenant selection problem. Both have a number of characteristics that make them an ideal testbed for the enhanced genetic algorithm using partnering strategies. Firstly, they are both in the class of NP-complete problems [Johnson 1998, Martello & Toth 1990]; hence, they are challenging problems. Secondly, they have proved resistant to optimisation by a standard genetic algorithm, with good solutions only found by using a novel strategy of indirectly optimising the problem with a decoder based genetic algorithm [Aickelin & Dowsland 2001]. Finally, both problems are similar multiple-choice allocation problems. For the nurse scheduling, the choice is to allocate a shift-pattern to each nurse, whilst for the tenant selection it is to allocate an area of the mall to a shop. However, as the following more detailed explanation of the two will show, the two problems also have some very distinct characteristics making them different yet similar enough for an interesting comparison of results.

The nurse-scheduling problem is that of creating weekly schedules for wards of up to 30 nurses at a major UK hospital. These schedules have to satisfy working contracts and meet the demand for given numbers of nurses of different grades on each shift, whilst at the same time being seen to be fair by the staff concerned. The latter objective is achieved by meeting as many of the nurses' requests as possible and by considering historical information to ensure that unsatisfied requests and unpopular shifts are evenly distributed. Due to various hospital policies, a nurse can normally only work a sub-set of the 411 theoretically possible shift-patterns. For instance, a nurse should work either days or nights in a given week, but not both. The interested reader is directed to Aickelin & Dowsland [2000] and Dowsland [1998] for further details of this problem.

For our purposes, the problem can be modelled as follows. Nurses are scheduled weekly on a ward basis such that they work a feasible pattern with regards to their contract and that the demand for all days and nights and for all qualification levels is covered. In total three qualification levels with corresponding demand exist. It is hospital policy that more qualified nurses are allowed to cover for less qualified one. Infeasible solutions with respect to cover are not acceptable. A solution to the problem would be a string, with the number of elements equal to the number of nurses. Each element would then indicate the shift-pattern worked by a particular nurse. Depending on the nurses' preferences, the recent history of patterns worked and the overall attractiveness of the pattern, a penalty cost is then allocated to each nurse-shift-pattern pair. These values were set in close consultation with the hospital and range from 0 (perfect) to 100 (unacceptable), with a bias to lower values. The sum of these values gives the quality of the schedule. 52 data sets are available, with an average problem size of 30 nurses per ward and up to 411 possible shift-patterns per nurse.

The problem can be formulated as an integer linear program as follows.

Indices:

$i = 1...n$ nurse index.

$j = 1...m$ shift pattern index.

$k = 1...7$ are days and $8...14$ are nights.

$s = 1...p$ grade index.

Decision variables:

$$x_{ij} = \begin{cases} 1 & \text{nurse } i \text{ works shift pattern } j \\ 0 & \text{else} \end{cases}$$

Parameters:

$n$ = Number of nurses.

$m$ = Number of shift patterns.

$p$ = Number of grades.

$$a_{jk} = \begin{cases} 1 & \text{shift pattern } j \text{ covers day / night } k \\ 0 & \text{else} \end{cases}$$

$$q_{is} = \begin{cases} 1 & \text{nurse } i \text{ is of grade } s \text{ or higher} \\ 0 & \text{else} \end{cases}$$

$p_{ij}$ = Preference cost of nurse $i$ working shift pattern $j$.

$N_i$ = Shifts per week of nurse $i$ if night shifts are worked.

$D_i$ = Shifts per week of nurse $i$ if day shifts are worked.

$B_i$ = Shifts per week of nurse $i$ if both are worked.

$R_{ks}$ = Demand of nurses with grade $s$ on day or night $k$.

$F(i)$ = Set of feasible shift patterns for nurse $i$, defined as

$$F(i) = \begin{cases} \sum_{k=1}^{7} a_{jk} = D_i & \forall j \in \text{day shifts} \\ \text{or} \\ \sum_{k=8}^{14} a_{jk} = N_i & \forall j \in \text{night shifts} \\ \text{or} \\ \sum_{k=1}^{14} a_{jk} = B_i & \forall j \in \text{combined shifts} \end{cases} \forall i$$

Target function:

$$\sum_{i=1}^{n} \sum_{j \in F(i)}^{m} p_{ij} x_{ij} \rightarrow \min!$$

Subject to:

$$\sum_{j \in F(i)} x_{ij} = 1 \qquad \forall i \qquad (1)$$

$$\sum_{j \in F(i)} \sum_{i=1}^{n} q_{is} a_{jk} x_{ij} \geq R_{ks} \quad \forall k,s \qquad (2)$$

Constraint set (1) ensures that every nurse works exactly one shift pattern from his/her feasible set, and constraint set (2) ensures that the demand for nurses is covered for every grade on every day and night. Note that the definition of $q_{is}$ is such that higher graded nurses can substituted those at lower grades if necessary. Typical problem dimensions are 30 nurses of three grades and 411 shift patterns. Thus, the Integer Programming formulation has about 12000 binary variables and 100 constraints.

Finally for all decoders, the fitness of completed solutions has to be calculated. Unfortunately, feasibility cannot be guaranteed, as otherwise an unlimited supply of nurses, respectively overtime, would be necessary. This is a problem-specific issue and cannot be changed. Therefore, we still need a penalty function approach. Since the chosen encoding automatically satisfies constraint set (1) of the integer programming formulation, we can use the following formula, where $w_{demand}$ is the penalty weight, to calculate the fitness of solutions. Hence the penalty is proportional to the number of uncovered shifts and the fitness of a solution is calculated as follows.

$$\sum_{i=1}^{n}\sum_{j=1}^{m} p_{ij} x_{ij} + w_{demand} \sum_{k=1}^{14}\sum_{s=1}^{p} \max\left[ R_{ks} - \sum_{i=1}^{n}\sum_{j=1}^{m} q_{is} a_{jk} x_{ij}; 0 \right] \to \min$$

Here we use an encoding that follows directly from the Integer Programming formulation. Each individual represents a full one-week schedule, i.e. it is a string of $n$ elements with $n$ being the number of nurses. The $ith$ element of the string is the index of the shift pattern worked by nurse $i$. For example, if we have 5 nurses, the string (1,17,56,67,3) represents the schedule in which nurse 1 works pattern 1, nurse 2 pattern 17 etc.

For comparison, all data sets were attempted using a standard Integer Programming package [Fuller 1998]. However, some remained unsolved after each being allowed 15 hours run-time on a Pentium II 200. Experiments with a number of descent methods using different neighbourhoods, and a standard simulated annealing implementation, were even less successful and frequently failed to find feasible solutions. A straightforward genetic algorithm approach failed to solve the problem [Aickelin & Dowsland 2000]. The best evolutionary results to date have been achieved with an indirect genetic approach employing a decoder function [Aickelin & Dowsland 2001]. However, we believe that there is further leverage in direct evolutionary approaches to this problem. Hence, we propose to use an enhanced pyramidal genetic algorithm in this paper.

## 3 TENANT SELECTION PROBLEM

The second problem is a mall layout and tenant selection problem; termed the mall problem here. The mall problem arises both in the planning phase of a new shopping centre and on completion when the type and number of shops occupying the mall has to be decided. To maximise revenue a good mixture of shops that is both heterogeneous and homogeneous has to be achieved. Due to the difficulty of obtaining real-life data because of confidentiality, the problem and data used in this research are constructed artificially, but closely modelled after the actual real-life problem as described for instance in Bean et al. [1988]. In the following, we will briefly outline our model.

The objective of the mall problem is to maximise the rent revenue of the mall. Although there is a small fixed rent per shop, a large part of a shop's rent depends on the sales revenue generated by it. Therefore, it is important to select the right number, size and type of tenants and to place them into the right locations to maximise revenue. As outlined in Bean et al. [1988], the rent of a shop depends on the following factors:

- The attractiveness of the area in which the shop is located.
- The total number of shops of the same type in the mall.
- The size of the shop.
- Possible synergy effects with neighbouring similar shops, i.e. shops in the same group (not used by Bean et al.).
- A fixed amount of rent based on the type of the shop and the area in which it is located.

This problem can be modelled as follows: Before placing shops, the mall is divided into a discrete number of locations, each big enough to hold the smallest shop size. Larger sizes can be created by placing a shop of the same type in adjacent locations. Hence, the problem is that of placing $i$ shop-types (e.g. menswear) into $j$ locations, where each shop-type can belong to one or more of $l$ groups (e.g. clothes shops) and each location is situated in one of $k$ areas. For each type of shop there will be a minimum, ideal and maximum number allowed in the mall, as consumers are drawn to a mall by a balance of variety and homogeneity of shops.

The size of shops is determined by how many locations they occupy within the same area. For the purpose of this study, shops are grouped into three size classes, namely small, medium and large, occupying one, two and three locations in one area of the mall respectively. For instance, if there are two locations to be filled with the same shop-type within one area, then this will be a shop of medium size. If there are five locations with the same shop-type assigned in the same area, then they will form one large and one medium shop etc. Usually, there will also be a maximum total number of small, medium and large shops allowed in the mall.

To test the robustness and performance of our algorithms thoroughly on this problem, 50 problem instances were created. All problem instances have 100 locations grouped into five areas. However, the sets differ in the number of shop-types available (between 50 and 20) and in the tightness of the constraints regarding the minimum and maximum number of shops of a certain type or size. Full details of the model and how the data was created, its dimensions and the differences between the sets can be found in [Aickelin 1999].

## 4 PYRAMIDAL GENETIC ALGORITHMS

Both problems failed to be optimised with a standard genetic algorithm [Aickelin & Dowsland 2000, 2001]. Our previous research showed that the difficulties were attributable to epistasis created by the constrained nature of the optimisation. Briefly, epistasis refers to the 'non-linearity' of the solution string [Davidor 1991], i.e. individual variable values which were good in their own right, e.g. a particular shift / location for a particular nurse / shop formed low quality solutions once combined. This effect was created by those constraints that could only be incorporated into the genetic algorithm via a penalty function approach. For instance, most nurses preferred working days; thus, partial solutions with many 'day' shift-patterns have a higher fitness. However, combining these shift-patterns leads to shortages at night and therefore infeasible solutions. The situation for the mall problem is similar yet more complex, as two types of constraints have to be dealt with: size constraints and number constraints.

In [Aickelin & Dowsland 2000] we presented a simple, and on its own unsuccessful, pyramidal genetic algorithm for the nurse-scheduling problem. A pyramidal approach can best be described as a hierarchical coevolutionary genetic algorithm where cascading clusters of sub-populations are built from bottom up. Higher-level sub-populations have individuals with loner strings and optimise larger parts of the problem. Thus, the hierarchy is not within one string but rather between sub-populations which optimise different problem portions. Hence, higher-level sub-populations search a larger search space with a lower resolution whilst lower-level sub-populations search a smaller search space with a higher resolution. A related hierarchical framework was presented using Genetic Programming [Koza 1991] whereby main program trees coevolve with successively lower level functions [e.g. Ahluwalia & Bull 1998]. The pyramidal GA can be applied to the nurse-scheduling problem in the following way:

- Solutions in sub-populations 1, 2 and 3 have their fitness based on cover and requests only for grade 1, 2 and 3 respectively.
- Solutions in sub-populations 4, 5 and 6 have their fitness based on cover and requests for grades (1+2), (2+3) and (3+1).
- Solutions in sub-population 7 optimise cover and requests for (1+2+3).
- Solutions in sub-population 8 solve the original (all) problem, i.e. cover for 1, for (1+2) and for (1+2+3).

The full structure is illustrated in figure 1. Sub-solution strings from lower populations are cascaded upwards using suitable crossover and selection mechanisms. For instance, fixed crossover points are used such that a solution from sub-population (1) combined with one from (1+2) forms a new solution in sub-population (1+2). Each sub-population performs 50% of crossovers uniform with two parents from itself. The other 50% are done by taking one parent from itself and the other from a suitable lower level population and then performing a fixed-point crossover. Bottom level sub-populations use only uniform crossover. The top level (all) population randomly chooses the second parent from all other populations. Although the full problem is as epistatic as before, the sub-problems are less so as the interaction between nurse grades is (partially) ignored. Compatibility problems of combining the parts are reduced by the pyramidal structure with its hierarchical and gradual combining. This can be seen as similar to the "Island Injection" parallel GA system [Eby et al. 1999].

Using this approach improved solution quality in comparison to a standard genetic algorithm was recorded. Initially roulette wheel selection based on fitness rank had been used to choose parents. The fitness of each sub-string is calculated using a substitute fitness measure based on the requests and cover as detailed above, i.e. the possibility of more qualified nurses covering for less-qualified ones is partially ignored. Unsatisfied constraints are still included via a penalty function. This paper will investigate various partnering strategies between the agents of the sub-populations to improve upon these results.

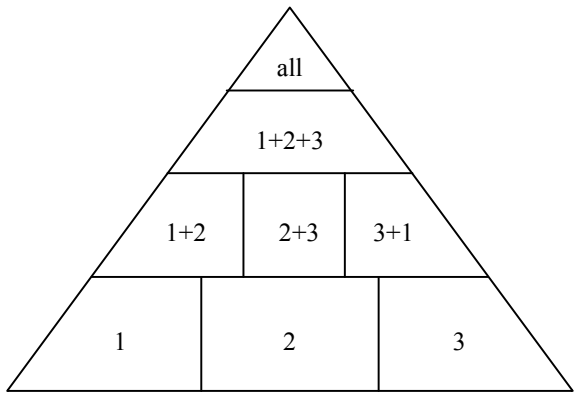

Figure 1: Nurse Problem Pyramidal Structure.

Similar to the nurse problem, a solution to the mall problem can be represented by a string with as many elements as locations in the mall. Each element then indicates what shop-type is to be located there. The mall is geographically split into different regions, for instance north, east, south, west and central. Some of the objectives are regional; e.g. the size of a shop, the synergy effects, the attractiveness of an area to a shop-type, whereas others are global, e.g. the total number of shops of a certain type or size.

The application of the pyramidal structure to the mall problem follows along similar lines to that of the nurse problem. In line with decomposing partitions into those with nurses of the same grade, the problem is now split into the areas of the mall. Thus, we will have sub-strings with all the shops in one area in them. These can then be combined to create larger 'parts' of the mall and finally full solutions.

However, the question arises how to calculate the substitute fitness measure of the partial strings. The solution chosen here will be a pseudo measure based on area dependant components only, i.e. global aspects are not taken into account when a substitute fitness for a partial string is calculated. Thus, sub-fitness will be a measure of the rent revenue created by parts of the mall, taking into account those constraints that are area based. All other constraints are ignored. A penalty function is used to account for unsatisfied constraints.

Due to the complexity of the fitness calculations and the limited overall population size, we refrained from using several levels in the hierarchical design as we did with the nurse scheduling. Instead a simpler two-level hierarchy is used as shown in figure 2: Five sub-populations optimising the five areas separately (1,2,3,4,5) and one main population optimising the original problem (all). Within the sub-populations 1-5 uniform crossover is used. The top-level population uses uniform crossover between two members of the population half the time and for the remainder a special crossover that selects one solution from a random sub-population that then performs a fixed-point crossover with a member of the top population.

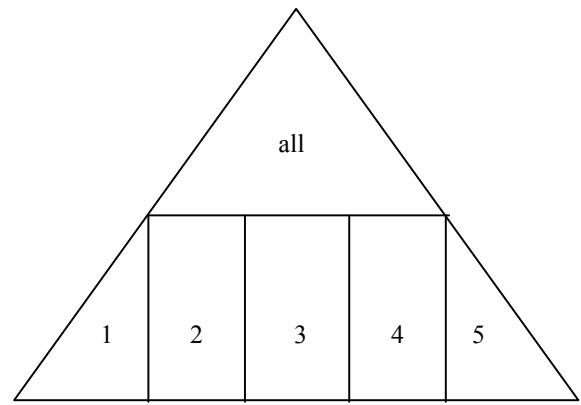

Figure 2: Mall Problem Pyramidal Structure.

The remainder of this paper will investigate ways to try to improve on previously found poor results by suggesting ways of combining partial strings more intelligently. An alternative, particularly for the mall problem, would be a more gradual build-up of sub-populations. Without increasing the overall population size, this would lead to more and hence smaller sub-populations. However, this more gradual approach might have enabled the algorithm to find good feasible solutions by more slowly joining together promising building blocks. This is in contrast to the relatively harsh two-level and three-level design where building blocks had to 'succeed' immediately. Exploring the exact benefits of a gradual build-up of sub-solutions would make for another challenging area of possible future research.

## 5   PARTNERING STRATEGIES

The problem of how to pick partners has been noted in both competitive and co-operative coevolutionary algorithms. Many strategies have been presented in the literature as summarised for instance in [Bull 1997]. In this paper, the following strategies are compared for their effectiveness in fighting epistasis by giving meaningful (sub-) fitness values in the pyramidal genetic algorithm optimising the nurse scheduling and the mall problems.

- Rank-Selection (S): This is the method used so far in our algorithms. Solutions are assigned a sub-fitness

score based as closely as possible on the contribution of their partial string to full solutions. All solutions are then ranked within each sub-population and selection follows a roulette wheel scheme based on the ranks [e.g. Aickelin & Dowsland 2000].

- Random (R): Solutions choose their mating partners randomly from amongst all those in the sub-population their sub-population is paired with [e.g. Bull & Fogarty 1993].

- Best (B): In this strategy, each agent is paired with the current best solution of the other sub-population(s). In case of a tie, the solution with the lower population index is chosen [e.g. Potter & De Jong 1994].

- Distributed (D): The idea behind this approach is to match solutions with similar ones to those paired with previously [e.g. Ackley & Littman 1994]. To achieve this each sub-population is spaced out evenly across a single toroidal grid. Subsequently, solutions are paired with others on the same grid location in the appropriate other sub-populations. Children created by this are inserted in an adjacent grid location. This is said to be beneficial to the search process because a consistent coevolutionary pressure emerges since all offspring appear in their parents' neighbourhoods [Husbands 1994]. In our algorithms, we use local mating with the neighbourhood set to the eight agents surrounding the chosen location.

- Best / Random (BR): A solution is paired twice: with the best of the other sub-population(s) and with a random partner(s). The better of the two fitness values is recorded.

- Rank-based / Random (SR): A solution is paired twice: with roulette wheel selected solution(s) and with (a) random partner(s). The better of the two fitness values is recorded.

- Random / Random (RR): A solution is paired twice with random partner(s). The better of the two fitness values is recorded.

# 6 EXPERIMENTAL RESULTS

## 6.1 THE MODEL

To allow for fair comparison, the parameters and strategies used for both problems are kept as similar as possible. Both have a total population of 1000 agents. These are split into sub-populations of size 100 for the lower-levels and a main population of size 300 for the nurse scheduling and respectively of size 500 for the mall problem. In principle, two types of crossover take place: within sub-populations a two-parent-two-children parameterised uniform crossover with p=0.66 for genes coming from one parent takes place.

Each new solution created undergoes mutation with a 1% bit mutation probability, where a mutation would re-initialise the bit in the feasible range. The algorithm is run in generational mode to accommodate the sub-population structure better. In every generation the worst 90% of parents of all sub-populations are replaced. For all fitness and sub-fitness function calculations a fitness score as described before is used. Constraint violations are penalised with a dynamic penalty parameter, which adjusts itself depending on the (sub)-fitness difference between the best and the best feasible agent in each (sub-)population. Full details on this type of weight and how it was calculated can be found in Smith & Tate [1993] and Aickelin & Dowsland [2000]. The stopping criterion is the top sub-population showing no improvement for 50 generations.

To obtain statistically sound results all experiments were conducted as 20 runs over all problem instances. All experiments were started with the same set of random seeds, i.e. with the same initial populations. The results are presented in feasibility and cost respectively rent format. Feasibility denotes the probability of finding a feasible solution averaged over all problem instances. Cost / Rent refer to the objective function value of the best feasible solution for each problem instance averaged over the number of instances for which at least one feasible solution was found.

Should the algorithm fail to find a single feasible solution for all 20 runs on one problem instance, a censored observation of one hundred in the nurse case and zero for the mall problem is made instead. As we are minimising the cost for the nurses and maximising the rent of the mall, this is equivalent to a very poor solution. For the nurse-scheduling problem, the cost represents the sum of unfulfilled nurses' requests and unfavourable shift-patterns worked. For the mall, the values for the rent are in thousands of pounds per year.

## 6.2 RESULTS

Table 1 shows the results for a variety of fitness evaluation strategies used and compares these to the theoretic bounds (Bound) and the standard genetic algorithm approach (SGA). For the Nurse Scheduling Problem all strategies used give better results than those found by the SGA. However, as explained above, most credit for this is attributed to the pyramidal structure reducing epistasis.

On closer examination, rank-based (S), random (R) and distributed (D) perform almost equally well, with the rank-based method being slightly better than the other two. All three methods have in common that they contain a stochastic element in the choice of partner. The benefit of this is apparent when compared to the best (B) method. Here the results are far worse which we attributed to the inherently restricted sampling. Interestingly, using the double schemes (SR, BR and RR) improves results across the board, which again strengthens our hypothesis how important good sampling is. The overall best results are found by the double random (RR) method. These results correspond to those reported in [Bull 1997].

The results for the Mall problem are similar to those found for the nurse problem: Double strategies work better than single ones and the Best strategy does particularly poorly. However, unlike for the nurse scheduling none of the single strategies significantly improves results over the SGA approach. Reasons for this have already been outlined in the previous sections, i.e. mainly the nature of splitting the problem into sub-problems being contrary to many of the problem's constraints. On the other hand, even for the simple strategies results are far improved over those found by using the partnering strategies for mating, whilst those found by the double strategies even outperform the SGA. We believe that this can be explained as follows: The main downfall of the partnering for mating strategies for the mall problem was outside control of these strategies. It lies in the fact that the sub-fitness scores are not a good predictor for the success of sub-solutions. However, as these results show, if the original (sub-)fitness measures are substituted by full fitness scores based on good partnering methods the pyramidal structure does work. This confirms our suspicion that the previous 'failure' of the pyramidal idea for the mall problem was rooted within our choice of sub-fitness measures rather than in the hierarchical sub-population idea itself.

| Method | N Cost | N Feasibility | M Rent | M Feasibility |
|--------|--------|---------------|--------|---------------|
| Bound  | 8.8    | 100%          | 2640   | 100%          |
| SGA    | 54.2   | 33%           | 1850   | 94%           |
| S      | 13.3   | 79%           | 1860   | 90%           |
| R      | 14.5   | 77%           | 1915   | 94%           |
| B      | 35.9   | 44%           | 1550   | 72%           |
| D      | 14.6   | 77%           | 1820   | 88%           |
| SR     | 12.7   | 84%           | 1950   | 99%           |
| BR     | 14.2   | 81%           | 1897   | 86%           |
| RR     | 12.1   | 83%           | 1955   | 99%           |

Table 1: Partnering Strategies for Fitness Evaluation Results (N = Nurse, M = Mall).

## 6.3 NURSE SCHEDULING WITH A HILLCLIMBER

The results presented so far show that even with the best algorithm for the nurse scheduling problem some data instances were unsolvable. In order to overcome this, a special hillclimber has been developed which is fully described in [Aickelin & Dowsland 2001]. The use of local search to refine solutions produced via the GA for complex problem domains is well established – often termed memetic algorithms [e.g. Moscato 1999]. Briefly, the hill-climber is local search based algorithm that iteratively tries to improve solutions by (chain-) swapping shift patterns between nurses or alternatively assigns a strictly solution improving pattern to a nurse. As the hill climber is computationally expensive, it is only used on those solutions showing favourable characteristics for it to exploit. Those solutions are referred to as 'balanced' and one example is a nurse surplus on one day shift and a shortage on another day shift.

The last set of experiments presented in table 2 shows what impact the best partnering schemes for evaluation (RR) has once the previously excluded hillclimber is attached to the genetic algorithm. The results reveal that the SGA is outperformed by the double random fitness evaluation approach coupled with the hill climber. One possible explanation for this effect can be found by having a closer look at the RR operator. Gains are most likely made due to better sampling. However, as mentioned before there is a large stochastic element involved in this case. Judging from these results it seems that this is beneficial as it leads to a bigger variety of solutions in turn leaving more for the hill climber to exploit.

| Algorithm | Short | N Cost | N Feasibility |
|-----------|-------|--------|---------------|
| SGA & Hillclimber | SGA&H | 10.8 | 91% |
| RR & Hillclimber | RR&H | 9.9 | 95% |

Table 2: Results for Algorithms combined with a Hillclimber for the Nurse Scheduling Problem.

## 7   CONCLUSIONS

Using the partnering strategies for evaluation purposes yields results in accordance with those reported in [Bull 1997]. For both problems the simple strategies worked equally well apart from the restricting 'best' choice. Combining two partnering schemes improved results further with the overall best solutions found by the double random strategy. Interestingly, the improvements of results seemed to be based on better sampling and more diversity. Thus for this approach an additional hillclimber is able to improve solutions beyond the previously best ones.